\documentclass[10pt,twocolumn,letterpaper]{article}

\usepackage[pagenumbers]{cvpr} 

\usepackage{graphicx}
\usepackage{amsmath}
\usepackage{amssymb}
\usepackage{booktabs}

\usepackage{mismath}
\usepackage{tabularray}
\usepackage{multirow}
\usepackage[noline]{algorithm2e}
\usepackage[hang,flushmargin]{footmisc}
\usepackage[accsupp]{axessibility} 

\usepackage[pagebackref,breaklinks,colorlinks]{hyperref}

\usepackage[capitalize]{cleveref}
\crefname{section}{Sec.}{Secs.}
\Crefname{section}{Section}{Sections}
\Crefname{table}{Table}{Tables}
\crefname{table}{Tab.}{Tabs.}

\def\cvprPaperID{42} 
\def\confName{ECV}
\def\confYear{2023}

\newcommand\manualfootnote[1]{%
  \begingroup
  \renewcommand\thefootnote{}\footnote{#1}%
  \addtocounter{footnote}{-1}%
  \endgroup
}

\begin{document}

\title{Vision Transformers with Mixed-Resolution Tokenization}

\author{
Tomer Ronen\\
Tel Aviv University\\
{\tt\small tomer.ronen34@gmail.com}
\and
Omer Levy\\
Tel Aviv University\\
\and
Avram Golbert\\
Google Research$^\text{*}$
}
\maketitle

\begin{abstract}
Vision Transformer models process input images by dividing them into a spatially regular grid of equal-size patches. Conversely, Transformers were originally introduced over natural language sequences, where each token represents a subword -- a chunk of raw data of arbitrary size. In this work, we apply this approach to Vision Transformers by introducing a novel image tokenization scheme, replacing the standard uniform grid with a mixed-resolution sequence of tokens, where each token represents a patch of arbitrary size. Using the Quadtree algorithm and a novel saliency scorer, we construct a patch mosaic where low-saliency areas of the image are processed in low resolution, routing more of the model's capacity to important image regions. Using the same architecture as vanilla ViTs, our Quadformer models achieve substantial accuracy gains on image classification when controlling for the computational budget. Code and models are publicly available at {\footnotesize \url{https://github.com/TomerRonen34/mixed-resolution-vit}}.
\end{abstract}
\section{Introduction}
Transformer~\cite{Vaswani2017AttentionIA} models are designed to process sequential input data. Vision Transformer (ViT)~\cite{Dosovitskiy2020AnII} models process input images that naturally have two spatial dimensions, requiring a spatially-aware tokenization scheme to convert them into sequences. The vast majority of Vision Transformers convert the input image into a two-dimensional grid of token vectors, before flattening it to create a one-dimensional sequence. Specifically, most methods use uniform patch tokenization, splitting the image into a spatially regular grid of equal-size patches.

In natural language processing, input tokenization looks entirely different. Almost all modern neural networks for text processing use subword tokenization, where each token represents a substring of arbitrary character length~\cite{Sennrich2015NeuralMT,Kudo2018SubwordRI}.
In this work, we apply this approach to ViTs by introducing a novel image tokenization scheme, replacing the standard uniform grid with a mixed-resolution sequence of tokens, where each token represents a patch of arbitrary size.

\begin{figure}[t!]
  \centering
  \includegraphics[width=0.82\linewidth]{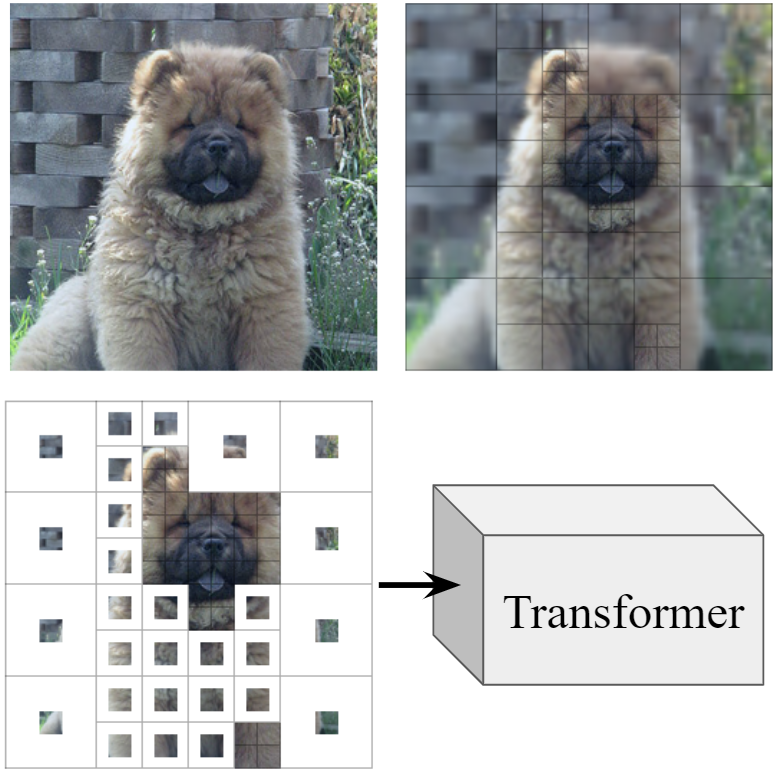}
  \caption{The Quadformer. We split the image into a mixed-resolution patch mosaic according to a saliency scorer, and employ a standard Transformer architecture with 2D position embeddings.}
  \label{figure:quadformer}
\end{figure}

Previous works tried to incorporate multi-resolution processing into Vision Transformers by building feature pyramids inspired by the structure of CNNs~\cite{Wang2021PyramidVT,Graham2021LeViTAV}, using multi-resolution attention~\cite{Yang2021FocalSF,Tang2022QuadTreeAF}, or merging intermediate token representations from across the entire image without preserving spatial locality~\cite{Renggli2022LearningTM,Bolya2022TokenMY}. In contrast, our work is the first to use mixed-resolution \textit{tokenization}, directly splitting the input image into a patch mosaic processed by a standard Transformer model (see Figure \ref{figure:quadformer}).

\manualfootnote{$^{\text{*}}$The author was affiliated with Alibaba Group during parts of the research.}

Instead of using a spatially regular patch grid, we construct a patch mosaic where low-saliency areas of the image are processed in low resolution, routing more of the model’s capacity to important areas. Practically, we use the Quadtree algorithm~\cite{Markas1992QuadTS} to recursively split the image into patches of different sizes, incorporating a saliency scorer that chooses which areas of the image to split by their estimated importance. We use 2D position embeddings to represent the location of each patch.

We evaluate our method, dubbed \textit{Quadformer}, on the ImageNet-1k~\cite{Russakovsky2014ImageNetLS} classification dataset, and compare our mixed-resolution models to vanilla ViT models that use the same architecture. While vanilla ViT models utilize uniform grid tokenizations with a single patch size (in our case, the standard $16^2$ pixels), our mixed-resolution tokenization uses 3 patch sizes ($64^2$, $32^2$ and $16^2$ pixels), allowing our Quadformer models to process important image regions in high resolution even when using a small number of patches. Using a novel saliency scorer based on neural representations, we consistently beat the accuracy of vanilla ViTs by up to 0.88 absolute percentage points when controlling for the number of patches or GMACs. Despite not using dedicated tools for accelerated inference, we also show gains when controlling for inference speed, beating vanilla ViT models by up to 0.42 absolute percentage points.

\section{Background and related work}

\paragraph{Efficient Vision Transformers.} Many efficient architectures were proposed for improving the speed-accuracy tradeoff of Vision Transformers, mostly by using attention layers with linear time complexity~\cite{Arar2021LearnedQF,Liu2021SwinTH,Tu2022MaxViTMV}, dropping a subset of patches~\cite{Meng2021AdaViTAV,Rao2021DynamicViTEV,Yin2021AViTAT}, or merging intermediate token representations from the entire image~\cite{Renggli2022LearningTM,Bolya2022TokenMY}. Our method offers orthogonal improvements as we decrease the number of patches via tokenization, maintaining global attention over the entire image while using spatially-local tokens.

\paragraph{Vision Transformers with spatially uniform grids.} Standard Vision Transformer models process input images by dividing them into a regular grid of equal size patches. Even in the case of pyramid vision transformers~\cite{Wang2021PyramidVT,Liu2021SwinTH}, which gradually compress the spatial dimension of the feature map as the network progresses, vectors in the same feature map always represent input areas of the same size. This is a classical design choice used extensively with CNNs, as it fits the constraints of convolution layers, that must operate on a spatially-regular grid. However, the layers that form the Transformer model, namely self-attention layers and fully connected layers, have no such limitations. Transformer models can process any set of input vectors that have some defined positional relationship, and are naturally suited to handling inputs of different scales. For example, Transformer language models process input tokens that represent subwords of very different lengths -- the BERT~\cite{Devlin2019BERTPO} vocabulary has tokens in lengths ranging from 1 character (``a'', ``b'') to 18 characters (``telecommunications'').

\begin{figure}[t!]
  \centering
  \vspace*{-5pt}
  \includegraphics[width=0.69\linewidth]{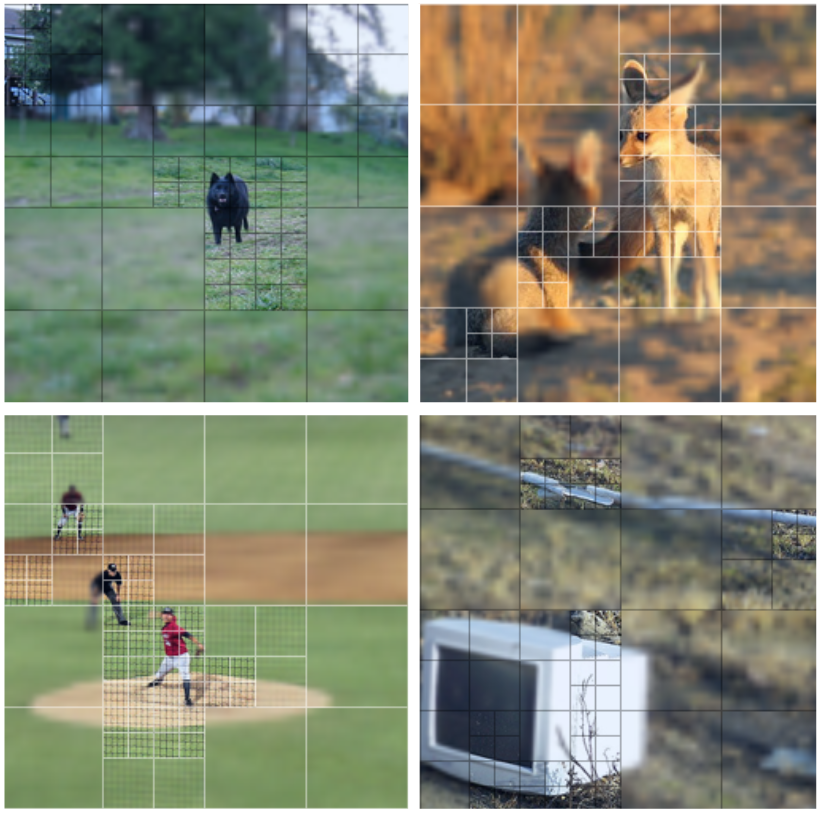}
  \caption{Tokenizations obtained using our saliency-based Quadtree. For clearer visualization, we upsample patches back to their original size after the tokenizer resizes them to a fixed representation size. Notice how high-saliency regions are represented in high resolution while background regions are blurry.}
  \label{figure:more_quadtree_examples}
\end{figure}

\paragraph{Existing methods for image tokenization.} Not all Vision Transformers use the standard uniform grid tokenization scheme. Some methods use CNN backbones to create representations from input images, using the activation volumes as tokens~\cite{Xiao2021EarlyCH,Graham2021LeViTAV}. Another class of Vision Transformers designed for image generation uses vector-quantization networks to learn a codebook of discrete tokens, also using a uniform two-dimensional grid~\cite{Esser2020TamingTF,Ramesh2021ZeroShotTG}. Few methods forgo spatial tokenization altogether and employ a technique called token learning, where each token aggregates information from the entire image~\cite{Ryoo2021TokenLearnerAS}.

\paragraph{Quadtrees.} Quadtrees are data structures that recursively split a two-dimensional space into a tree of quadrants, where each internal node has exactly four children. Each node in the tree represents a specific spatial area defined by an axis-aligned rectangle or square. Leaf nodes store the information contained in the area they represent. Quadtrees were originally developed for fast retrieval of 2D points~\cite{Finkel1974QuadTA}. They were quickly adapted for image analysis~\cite{Hunter1979OperationsOI}, and later for image compression~\cite{Markas1992QuadTS}.

\begin{figure*}[t!]
  \vspace*{-5pt}
  \centering
  \includegraphics[width=0.7\linewidth]{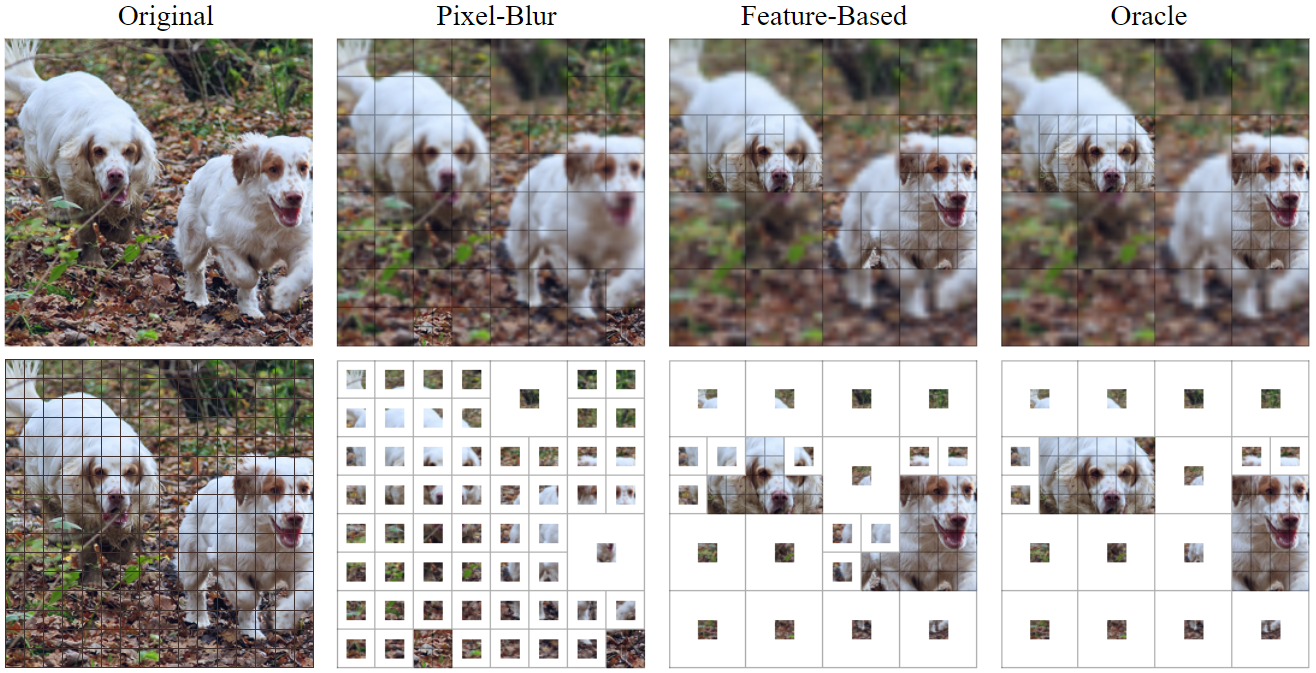}
  \caption{The effect of different patch scorers on Quadtree tokenization. Better saliency estimator $\rightarrow$ higher resolution in important areas. The pixel-blur scorer is often used for image compression, as it focuses on high-frequency details. Our feature-based scorer estimates patch saliency using neural representations. The oracle scorer uses the Grad-CAM saliency estimation algorithm.}
  \label{figure:quadtree_comparison}
\end{figure*}

\paragraph{Quadtrees and neural networks.} Few successful attempts have been made to integrate the Quadtree algorithm with neural networks. To the best of our knowledge, our work is the first to use Quadtree representations of RGB images as inputs to a neural net.

Jewsbury \etal~\cite{Jewsbury2021AQI} use Quadtrees to divide large pathology images into smaller subimages, with each subimage individually processed by a standard CNN. Many works on 3D shape analysis~\cite{Riegler2016OctNetLD,Tatarchenko2017OctreeGN,Wang2017OCNN} use specialized CNN architectures to process Octrees~\cite{Meagher1980Octree}, the 3D equivalent of Quadtrees. Jayaraman \etal~\cite{Jayaraman2018QuadtreeCN} use Quadtrees with sparse CNNs to process simple black-and-white sketches, avoiding computation in blank areas of the image. Chitta \etal~\cite{Chitta2019QuadtreeGN} use Quadtrees with a sparse CNN decoder to predict hierarchical segmentation maps, avoiding excessive computation in large image regions that share the same class. Tang \etal~\cite{Tang2022QuadTreeAF} propose an efficient attention implementation for ViTs, where each query vector in a spatially uniform grid attends to a Quadtree of key-value vectors. Ke \etal~\cite{Ke2021MaskTF} use Quadtrees for efficient refinement of instance segmentation masks, focusing computation in incoherent regions. They employ a Transformer model over a Quadtree of feature vectors extracted from a CNN feature pyramid.


\section{Method}

\subsection{ViTs with mixed-resolution tokenization}

We define a \textbf{mixed-resolution patch mosaic} to be a division of an image into a set of non-overlapping patches of different sizes, such that the entire area of the image is covered (see examples in Figure \ref{figure:quadformer}, Figure \ref{figure:more_quadtree_examples}, Figure \ref{figure:quadtree_comparison}). With small adaptions to the way ViT models represent image patches, we convert mixed-resolution patch mosaics into token sequences that can be processed by a standard Transformer model. These adaptations deal with 2 aspects of the tokens: patch embedding and position embedding.

\noindent \textbf{Patch embedding:} each patch in the mosaic is resized to a fixed representation size (e.g. $16^2$ pixels), then flattened and passed through a shared fully connected layer. Notice that all patches are represented by tokens of equal dimension, regardless of the area they cover in the image.

\noindent \textbf{Position embedding:} the learned 1-dimensional position embeddings common in vanilla ViTs lose meaning when the patches are not part of a regular grid. Instead, we use 2-dimensional position embeddings. We embed the $x$ and $y$ positions separately, then concatenate them to create the final position embedding, as suggested by Dosovitskiy \etal~\cite{Dosovitskiy2020AnII} We use the $(x,y)$ position of the center of the patch inside a grid determined by the smallest patch size.

\begin{algorithm}[t!]
\footnotesize
\textbf{Input:}

Image~~$im\in\mathbb{R}^{h \times w \times 3}$~,

\vspace*{1pt}
desired number of patches $L \in \mathbb{N}$~,

\vspace*{1pt}
patch edge sizes~~$s_{min},s_{max} \in \mathbb{N}$~,

\vspace*{1pt}
saliency scorer~~$score: patch \to \mathbb{R}^+$

\vspace*{1pt}
\textbf{Output:}

The set of chosen patches $P_{chosen}$

\vspace*{6pt}
\textbf{Algorithm:}

\vspace*{1pt}
$P_{chosen} \gets$ slice $im$ into a uniform grid with patch size $s_{max}$

\vspace*{3pt}
\While{$~\lvert P_{chosen} \lvert~<~L~$}{

\vspace*{1pt}
$P_{splittable} \gets \{p~~|~~p \in P_{chosen}~~\&~~size(p) \ge 2s_{min}\}$

\vspace*{1pt}
$p_{split} \gets \arg\max_{p \in P_{splittable}} score(p) $

\vspace*{2pt}
$children(p_{split}) \gets$ divide $p_{split}$ into 4 quadrants

\vspace*{1pt}
$P_{chosen} \gets children(p_{split}) \cup P_{chosen} \setminus \{ p_{split} \} $
}

\vspace*{3pt}

\textbf{Return} $P_{chosen}$

\vspace*{10pt}

\caption{\small The saliency-based Quadtree. We iteratively choose the ``most important'' image region as ranked by a saliency scorer and split it into 4 quadrants. In practice, we run the algorithm on a batch of images for improved speed, taking only $19 \mu\text{-}secs$ per image for the splitting logic. Patch scoring is also batched, taking $19\text{--}157 \mu\text{-}secs$ per image depending on the scorer. See subsection \ref{paragraph:impl_quadtree} and Table \ref{table:cost_breakdown} for more details.}
\label{algorithm:quadtree}
\end{algorithm}

\subsection{Saliency-based Quadtrees}

\paragraph{Quadtrees for RGB images.} Quadtrees are data structures that recursively split a two-dimensional space into a tree of quadrants, where each internal node has exactly four children. Each node in the tree represents a specific spatial area defined by an axis-aligned rectangle or square. In Quadtrees that represent RGB images, each leaf contains a compressed representation of an image patch, often a copy of that patch downsampled to some predetermined size.

Typically, Quadtrees for RGB images are constructed by a top-down algorithm (Algorithm \ref{algorithm:quadtree}), which iteratively chooses the ``most important'' image patch as ranked by a scoring function and splits it into 4 patches, effectively using 4 times more pixels to represent the selected image region. We call this scoring function a ``patch scorer''.

We use the Quadtree algorithm as a tokenizer, splitting images into mixed-resolution patch mosaics which we then feed into a standard Transformer model. We experiment with several patch scorers (Figure \ref{figure:quadtree_comparison}): the pixel-blur scorer commonly used for Quadtree image compression, a novel feature-based scorer that estimates saliency using neural representations, and a Grad-CAM oracle scorer which utilizes a label-aware saliency method and gives a loose upper bound on the scoring quality we can hope to achieve.

\begin{figure}[t!]
  \centering
  \vspace*{-12pt}
  \includegraphics[width=0.52\linewidth]{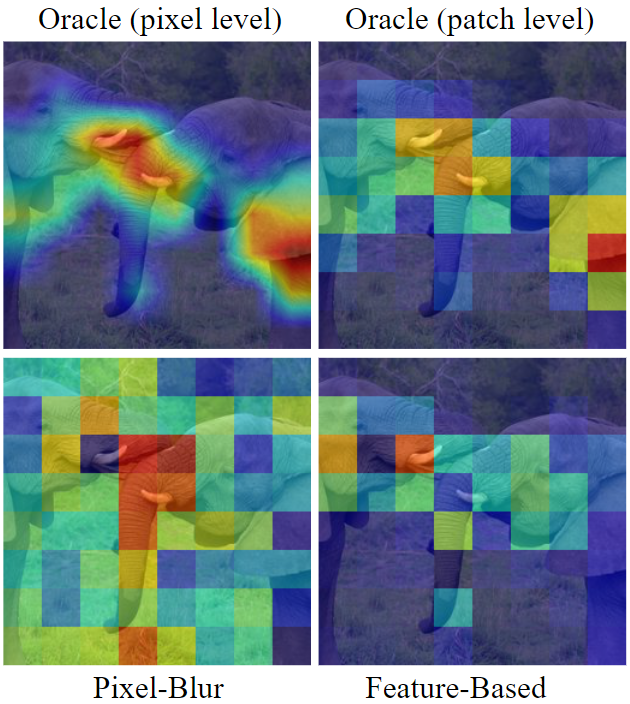}
  \caption{Patch saliency maps created by different scorers for an image labeled ``African Elephant''.}
  \label{figure:importance_map_comparison}
\end{figure}

\paragraph{Pixel-blur scorer.} In image compression applications, Quadtree patch scoring often relies on the MSE between an image patch and a compressed representation of that patch~\cite{Markas1992QuadTS}, such as a blurry version of the patch obtained by downsampling it to the Quadtree representation size and upsampling back to the original size. This score estimates the pixel-level information loss caused by decreasing the resolution of the patch. Let $p$ be an image patch:
\begin{align}
\begin{aligned}
&p_{blur} = upsample \bigl( downsample(p) \bigr) \\
&score_{PixelBlur}(p) = MSE \bigl( p,\ p_{blur} \bigr)
\end{aligned}
\label{equation:pixel_blur_scorer}
\end{align}

The pixel blur scorer assigns high importance to areas of the image with a lot of high-frequency content, since calculating the difference between a patch and its blurry counterpart is equivalent to running a high-pass filter. While high-frequency content may be a good importance measure for image compression, it is a poor measure of object saliency, as natural images often have detailed backgrounds or textures that are insignificant when trying to identify the objects in the image. To address this misalignment between the patch scorer objective and the model objective, we propose a different scorer based on semantic representations.

\begin{figure}[t!]
  \centering
  \vspace*{-10pt}
  \includegraphics[width=0.8\linewidth]{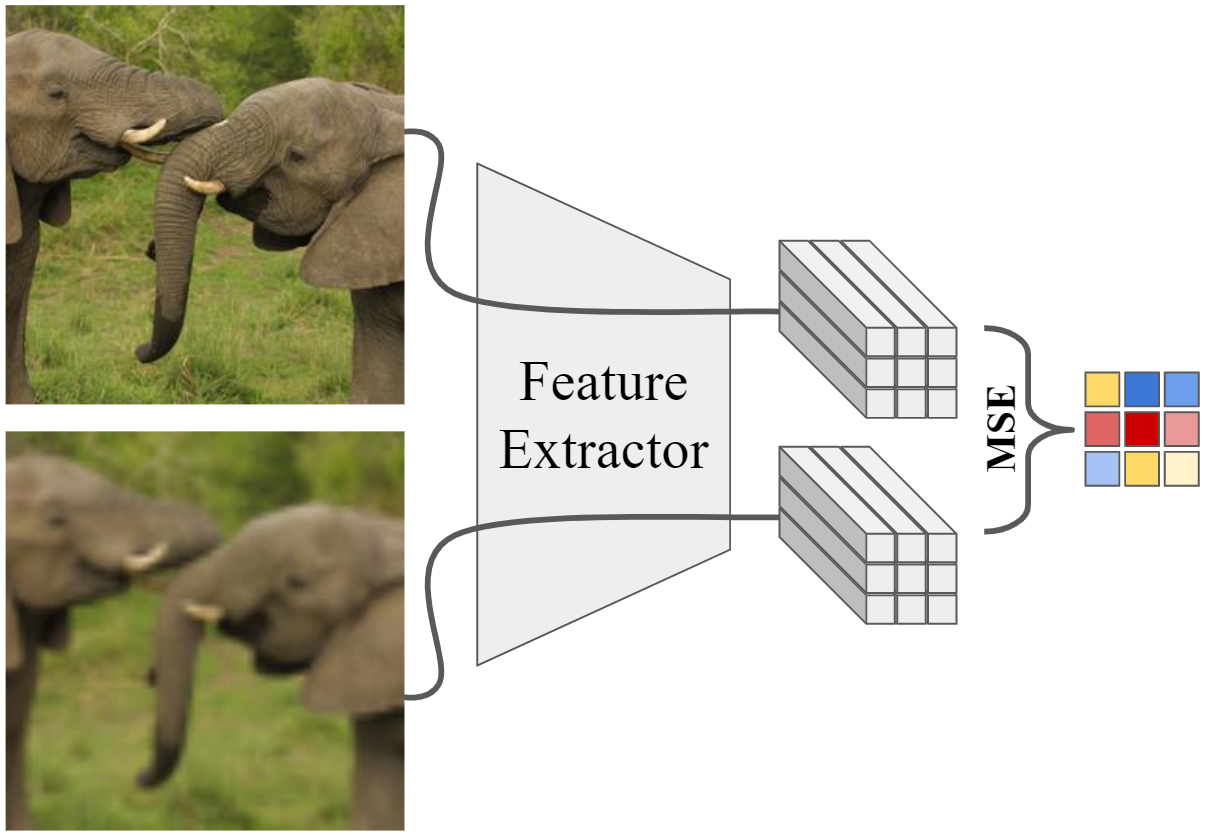}
  \caption{Feature-based patch scorer. The MSE between a patch representation and its blurry counterpart estimates the semantic information loss from decreasing the resolution of the patch.}
  \label{figure:feature_based_scorer}
\end{figure}

\paragraph{Feature-based scorer.}\label{paragraph:feature_based_scorer}
Computer vision neural networks are often used to extract semantically meaningful feature vectors. Both Vision Transformers and CNNs create contextualized embeddings of image regions: ViTs have an explicit mapping between feature vectors to image patches, and CNNs create a spatially-aware convolutional activation volume for the entire image wherein each feature vector can be mapped implicitly to a corresponding image region.

Using these neural representations, we introduce $score_{Feat}$, a patch scorer that estimates the semantic information loss from decreasing the resolution of an image patch by comparing its original representation to its representation in a blurred image (Figure \ref{figure:feature_based_scorer}). Intuitively, this score estimates how much semantic information is lost when we downsample the patch from its original size to the Quadtree representation size. For example, if the Quadtree representation size is $16^2$ pixels, the features of a $64^2$ patch in full resolution are compared to the features of this patch when the image is blurred by a factor of $\frac{64}{16}=4$.

Formally,  let $im\in\mathbb{R}^{h_{im} \times w_{im} \times 3}$ and $blur(im, x)$ be an RGB image and its corresponding blurred image obtained by downsampling the image by a factor of $x$ and upsampling it back to the original size. We extract a feature map $feat(im)\in\mathbb{R}^{H \times W \times d}$ by running a feature extractor NN on the image $im$. Given an image patch $p$ of size $s_p \times s_p$, we slice the region in $feat(im)$ which corresponds to $p$'s location in the image: $feat(im)[p]\in\mathbb{R}^{\frac{s_p}{h_{im}}H \times \frac{s_p}{w_{im}}W \times d}$. We use $feat(im)[p]$ as a semantic representation of $p$, a technique very similar to RoI pooling~\cite{Girshick2015FastR}. Given the Quadtree representation size $s_{rep} \in \mathbb{N}$, we use these notations to \linebreak define $score_{Feat}$:
\begin{align}
\begin{aligned}
&im_{blur} = blur \bigl( im,\frac{s_p}{s_{rep}} \bigr)  \\
&score_{Feat}(p) = MSE \bigl( feat(im_{blur})[p],~feat(im)[p] \bigr)
\end{aligned}
\label{equation:feature_base_scorer}
\end{align}

\paragraph{Grad-CAM oracle scorer.}\label{paragraph:oracle_scorer}
Grad-CAM~\cite{Selvaraju2016GradCAMVE} is a method for creating visual explanations of predictions made by a variety of computer vision models. For classification nets, given an image and a target class, Grad-CAM produces a pixel-level saliency map where the weight attributed to each pixel represents its importance in classifying the image to the given target class. Using average pooling, we turn this saliency map into patch scores suitable for the saliency-based Quadtree algorithm (Figure \ref{figure:importance_map_comparison}). To estimate a loose upper bound on the accuracy we can hope to achieve with Quadformer models, we use specific oracle Quadformers, which we train and evaluate with the high-quality saliency scores produced by a Grad-CAM patch scorer that is aware of the actual ground-truth label of the input images.

\section{Experiments}

\subsection{Dataset and evaluation metrics}
We conduct experiments on ImageNet-1K~\cite{Russakovsky2014ImageNetLS} and report the top-1 accuracy trade-off with respect to several cost indicators, as suggested by Dehghani \etal~\cite{Dehghani2021TheEM} To evaluate model efficiency, we report the number of patches/tokens in the input to the Transformer model, the number of giga multiply--accumulate operations (GMACs) per image as estimated by fvcore~\cite{fvcore_flop_counting}, and the throughput (ims/sec) and runtime ($\mu$-secs/im) on a single GeForce RTX 3090 GPU, measured with timm~\cite{rw2019timm} with batch size 512 in mixed precision. We do not use parameter count as a cost indicator since our Quadformer models use the exact same architectures as our vanilla ViT models: ViT-Small (22M params), ViT-Base (86M params) and ViT-Large (307M params). Some Quadformer models use a neural net for saliency estimation, but since it only has 342K parameters (see \S\ref{paragraph:impl_patch_scorers}) its impact on the parameter count is negligible.

\subsection{Implementation details}

\paragraph{Base models.}
All our base models use image size $256^2$, patch size $16^2$, and $2D$ sinusoidal position embeddings. For our two main ViT architectures --- ViT-Base and ViT-Large --- we start by taking the weights released by the original authors~\cite{Dosovitskiy2020AnII}, which are pretrained on ImageNet-21K and fine-tuned on ImageNet-1K. These pretrained models use learned 1D position embeddings, image size $224^2$, and patch size $16^2$. We adapt them to $2D$ sinusoidal position embeddings and image size $256^2$ by fine-tuning on ImageNet-1K with base learning rate $1\text{e-}4$ for 70 epochs (for ViT-Base) or 20 epochs (for ViT-Large). For each architecture, we choose the checkpoint that achieved the highest validation accuracy. For ViT-Small, we train the DeiT-S architecture~\cite{Touvron2020TrainingDI} from scratch on ImageNet-1K with base learning rate $2\text{e-}3$ for 310 epochs.

\paragraph{Fine-tuning.}
We use the base models to initialize the weights of all our fine-tuned models, as we have seen much faster conversion times compared to training from scratch. We use the same base models to initialize both vanilla Vision Transformers and Quadformer models, as Quadformers share the exact same architecture with vanilla ViTs and do not introduce any extra parameters, except those used in the tokenizer.

Our Quadformer models use mixed-resolution tokenizations with patch sizes $64^2$, $32^2$ and $16^2$ pixels, all downsampled to a patch representation size of $16^2$ pixels. We fix the image size to $256^2$ pixels and control the number of patches by setting the number of splits done by the Quadtree algorithm. Our vanilla ViT models use patch size $16^2$. We control the number of patches by setting the image size to $(16\sqrt{\#Patches})^2$ pixels. We report detailed hyperparameters in the supplementary material.

\paragraph{Patch scorers.} \label{paragraph:impl_patch_scorers}
For our feature-based patch scorer, we use a ShuffleNetV2$\times0.5$~\cite{Ma2018ShuffleNetVP} model trained on ImageNet-1K as the feature extractor. We truncate it just before the fully connected classification layer, which results in a $\times32$ downscaling ratio. This feature extraction backbone has only 342K parameters\label{342k_params}, which adds little overhead and makes it practical for real-world inference purposes. For Quadformer models that use ViT-Base or ViT-Small, we perform the scoring on a $\times0.75$ downsampled image (with $192^2$ pixels) and then upsample the saliency map by the same ratio, since the increased speed compensates for the lower fidelity and results in a better speed-accuracy tradeoff.

For Grad-CAM oracle saliency estimation we use a RegNetY-32GF~\cite{Radosavovic2020DesigningND} model with 145M parameters, learned via transfer learning by end-to-end fine-tuning the original SWAG~\cite{Singh2022RevisitingWS} weights on ImageNet-1K data. Weights for ShuffleNetV2$\times0.5$ and RegNetY-32GF are taken from the torchvision library~\cite{torchvision2016}.

\paragraph{Quadtree.} \label{paragraph:impl_quadtree}
We build our own PyTorch~\cite{Paszke_PyTorch_An_Imperative_2019} implementation of the Quadtree algorithm (Algorithm \ref{algorithm:quadtree}), using z-order curves for efficient tree construction~\cite{Warren1993APH}. Patch scores are computed for an entire batch of input images over all possible spatial locations. Since we use a top-down algorithm, the only valid candidates are subdivisions of the initial patch grid, resulting in a total of 80 splittable patches for images of size $256^2$ pixels. All our patch scorers employ image-wide computation followed by grid-based scoring, making them particularly suitable for this kind of batched computation. The argmax operation used for iterative splitting is also batched, as well as the image slicing and resizing required to create token representations, making the entire implementation very GPU-friendly.

\subsection{Main results}
\begin{figure}[t!]
  \centering
  \hspace*{-0.035\linewidth}
  \includegraphics[width=1.035\linewidth]{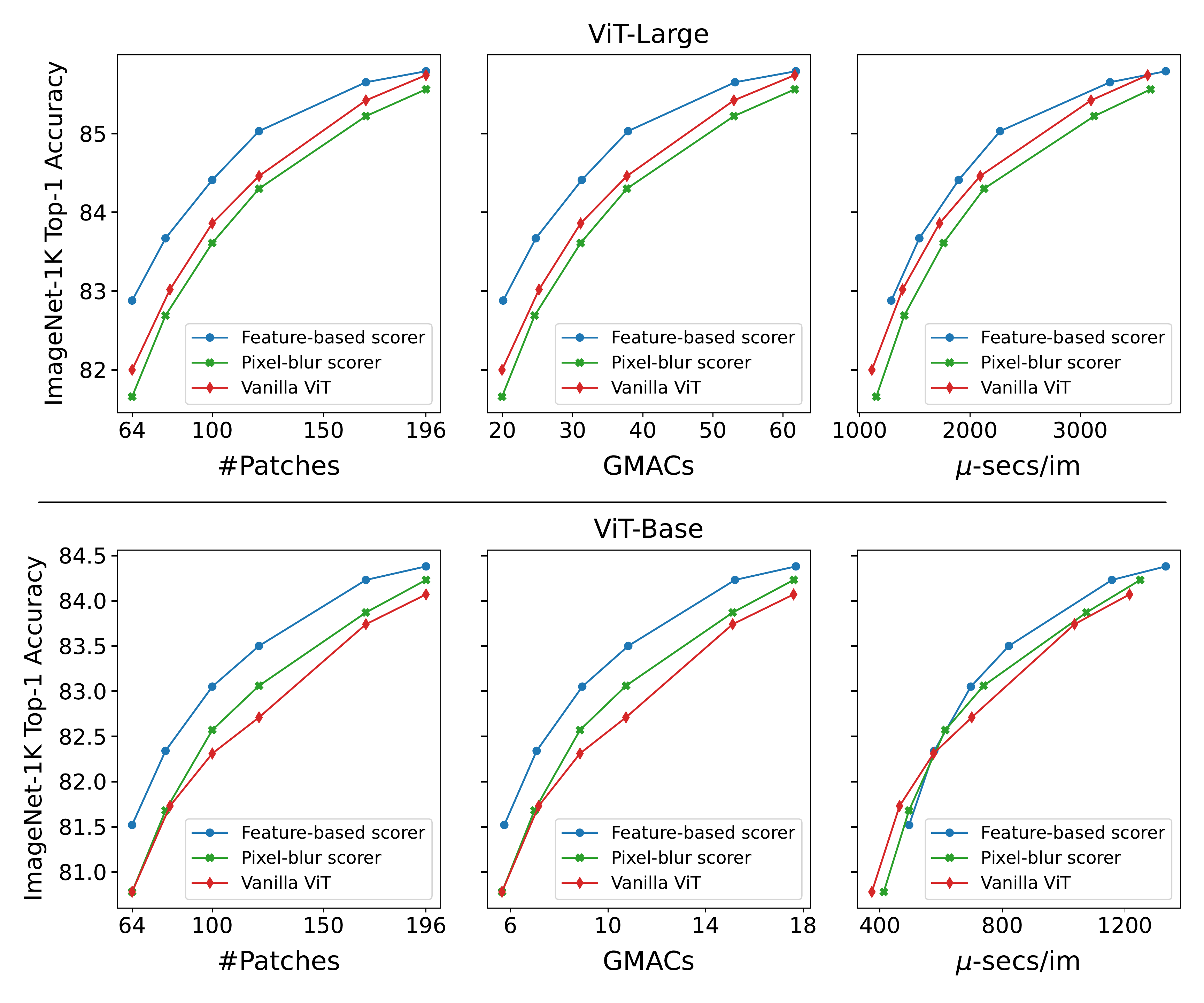}
  \caption{Accuracy vs compute for vanilla ViT and Quadformer models with different saliency scorers. Every point represents a model fine-tuned with a specific number of patches. We expect the performance of Quadformer and vanilla models to converge as \#Patches approaches full resolution (256 patches). Throughput is measured on a single GeForce RTX 3090 GPU in mixed precision.}
  \label{figure:accuracy_vs_compute_main_results}
\end{figure}

Using a feature-based scorer (\S\ref{paragraph:feature_based_scorer}), our Quadformer models consistently beat the accuracy of vanilla Vision Transformers by up to 0.79 (for ViT-Base) or 0.88 (for ViT-Large) absolute percentage points when controlling for the number of patches or GMACs, while using the exact same architecture (see Figure \ref{figure:accuracy_vs_compute_main_results}). Despite not using dedicated tools for accelerated inference, we also show gains when controlling for inference speed, beating vanilla ViT models for almost all values of \#Patches by up to 0.42 (for ViT-Base) or 0.4 (for ViT-Large) absolute percentage points. The traditional pixel-based scorer used for image compression fairs much worse than our feature-based scorer, demonstrating the superiority of semantic meaning over surface details. Full results are provided in the supplementary material.

\subsection{Inference-time compute-accuracy tradeoff}
Both Quadformers and vanilla Vision Transformers can be trained with a certain number of patches and operate on inputs with a different number of patches, providing a way to control the compute-accuracy tradeoff of a single model during inference time. With Quadformers, we use a different number of Quadtree splits to produce tokenizations of different lengths, allowing high granularity as every split increases the number of patches by 3 -- the split patch is replaced with its 4 children patches. With vanilla ViTs, we change the image size to a different multiple of the patch size, thus changing the total number of patches. When using vanilla ViTs with different image sizes, we scale the 2D patch positions to fit the range seen during training time, as we have seen better results when the inference-time position embeddings closely resemble those seen while training.

\begin{figure}[t!]
  \centering
  \vspace*{-4pt}
  \hspace*{-0.02\linewidth}
  \includegraphics[width=1.0\linewidth]{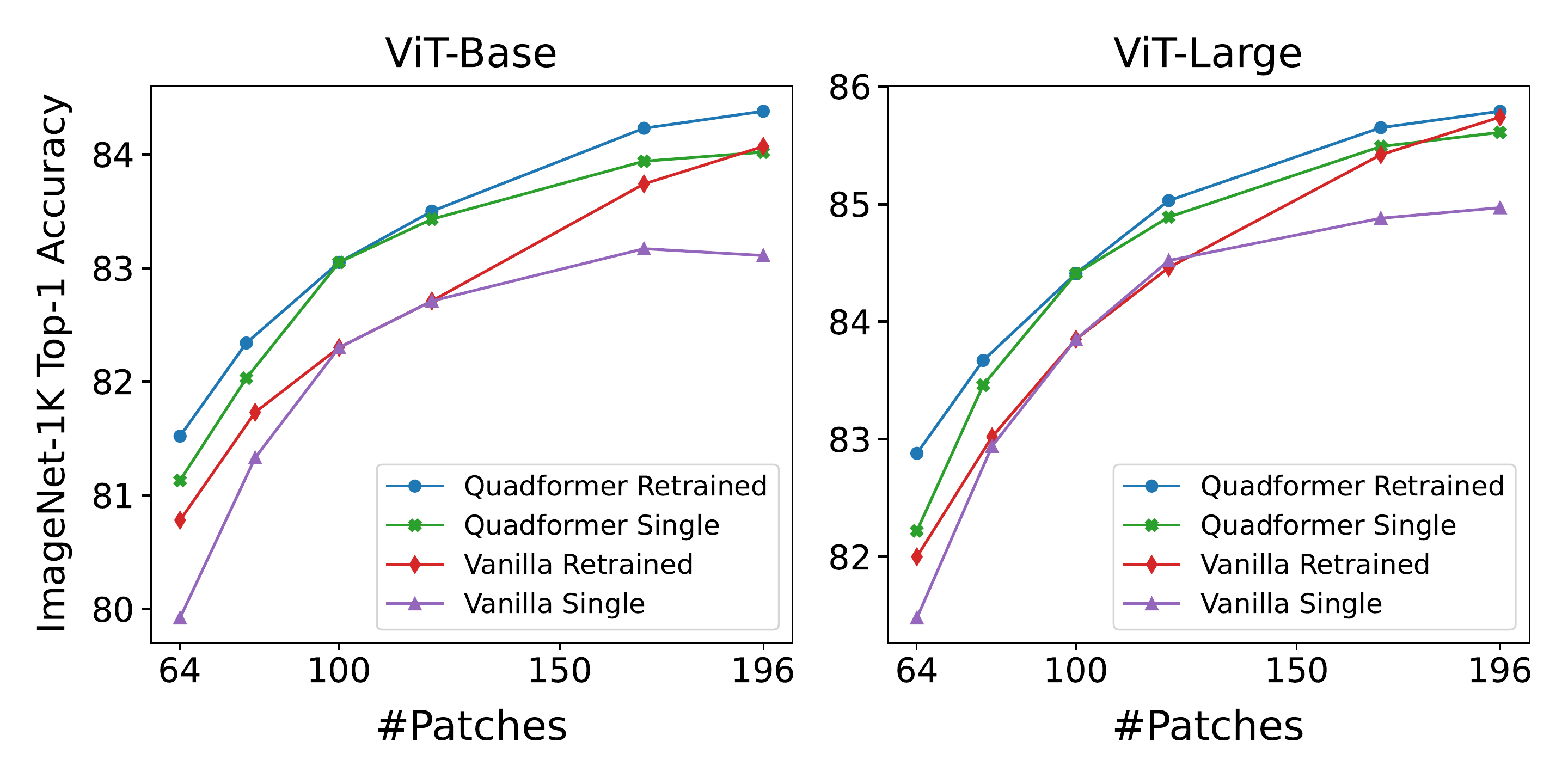}
  \caption{Inference-time compute-accuracy tradeoff for Quadformer models with a feature-based scorer and vanilla ViTs. ``Retrained'' lines show models that are retrained for each value of \#Patches. ``Single'' lines show a single model (trained with 100 patches) evaluated with different \#Patches. Quadformers are less sensitive to out-of-distribution input lengths, providing a better inference-time compute-accuracy tradeoff with a single model.}
  \label{figure:accuracy_vs_compute_single_model_comparison}
\end{figure}

Figure \ref{figure:accuracy_vs_compute_single_model_comparison} compares the inference-time compute-accuracy tradeoff of a single Quadformer model with a feature-based scorer and a single vanilla ViT model to versions of these models specifically trained for each number of patches. Quadformers are less sensitive to out-of-distribution input lengths, showing a lower accuracy drop with respect to their retrained counterparts, and providing a better inference-time compute-accuracy tradeoff with a single model.

\subsection{Small Quadformers}
\begin{figure}[t!]
\vspace*{-8pt}
\centering
\hspace*{-0.035\linewidth}
\includegraphics[width=1.045\linewidth]{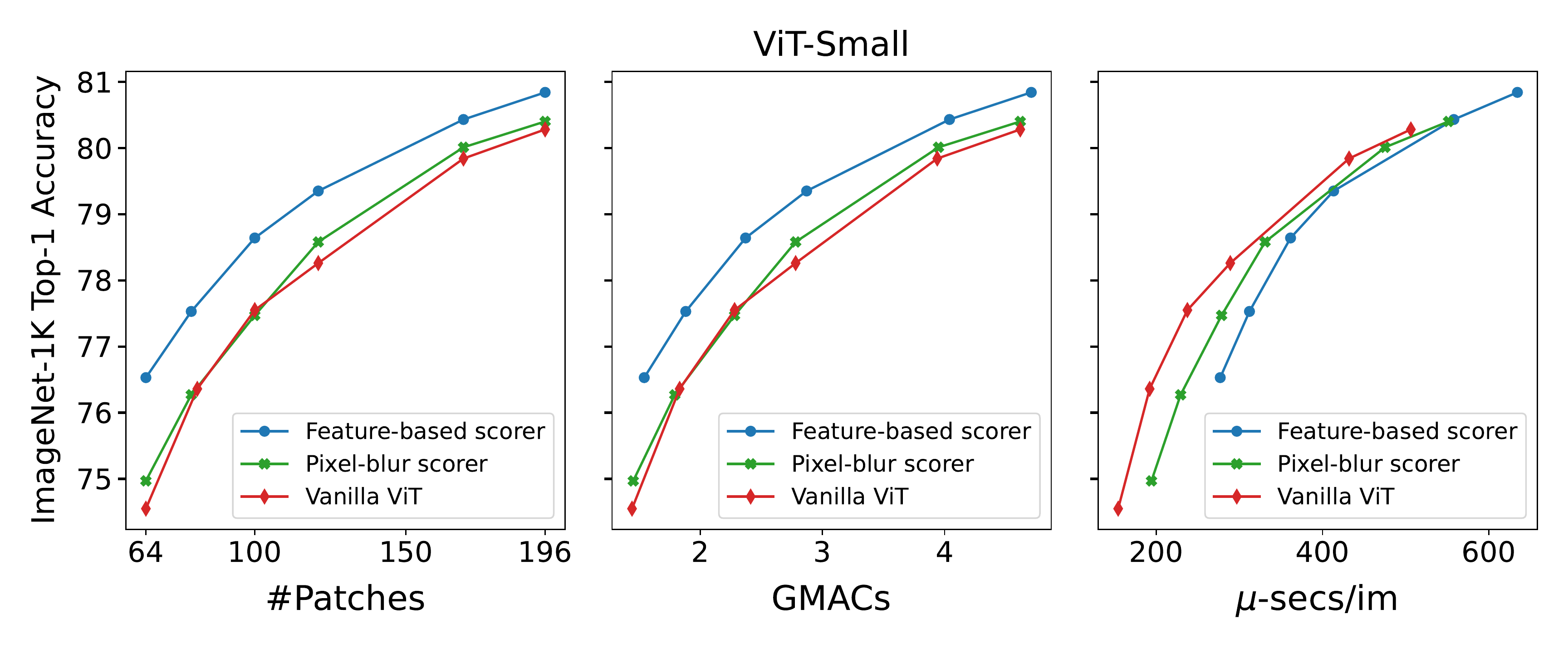}
\caption{Accuracy vs compute for vanilla ViT-Small and Quadformer-Small models with different saliency scorers. Small Transformers pose an interesting challenge, being so fast that any tokenization overhead is significant.}
\label{figure:accuracy_vs_compute_small}
\end{figure}

Small Transformers pose an interesting challenge. On the one hand, weak models have the most to gain from high-quality saliency estimation, since they lack the capacity required to compensate for low-resolution images or mediocre patch selection. Quadformer-Small beats the accuracy of vanilla ViT-Small by up to $1.98$ absolute percentage points when controlling for the number of patches, and by up to $1.54$ points when controlling for GMACs. On the other hand, small Transformers are so fast that the runtime of the feature-based scorer is too costly compared to the total runtime (Figure \ref{figure:runtime_breakdown}) making it inefficient in terms of runtime-accuracy tradeoff, even compared to the weak, yet speedy, pixel-blur scorer (Figure \ref{figure:accuracy_vs_compute_small}).

Future work may find faster high-quality saliency estimators that would enable small Vision Transformers to use mixed-resolution tokenization efficiently. We note that many previous works dealing with efficient Vision Transformers~\cite{Arar2021LearnedQF,Liu2021SwinTH,Tu2022MaxViTMV,Meng2021AdaViTAV} do not report results for models that are as fast as ViT-Small, perhaps encountering similar issues with speeding up such fast models.

\begin{figure}[t!]
\vspace*{-2pt}
\centering
\includegraphics[width=0.9\linewidth]{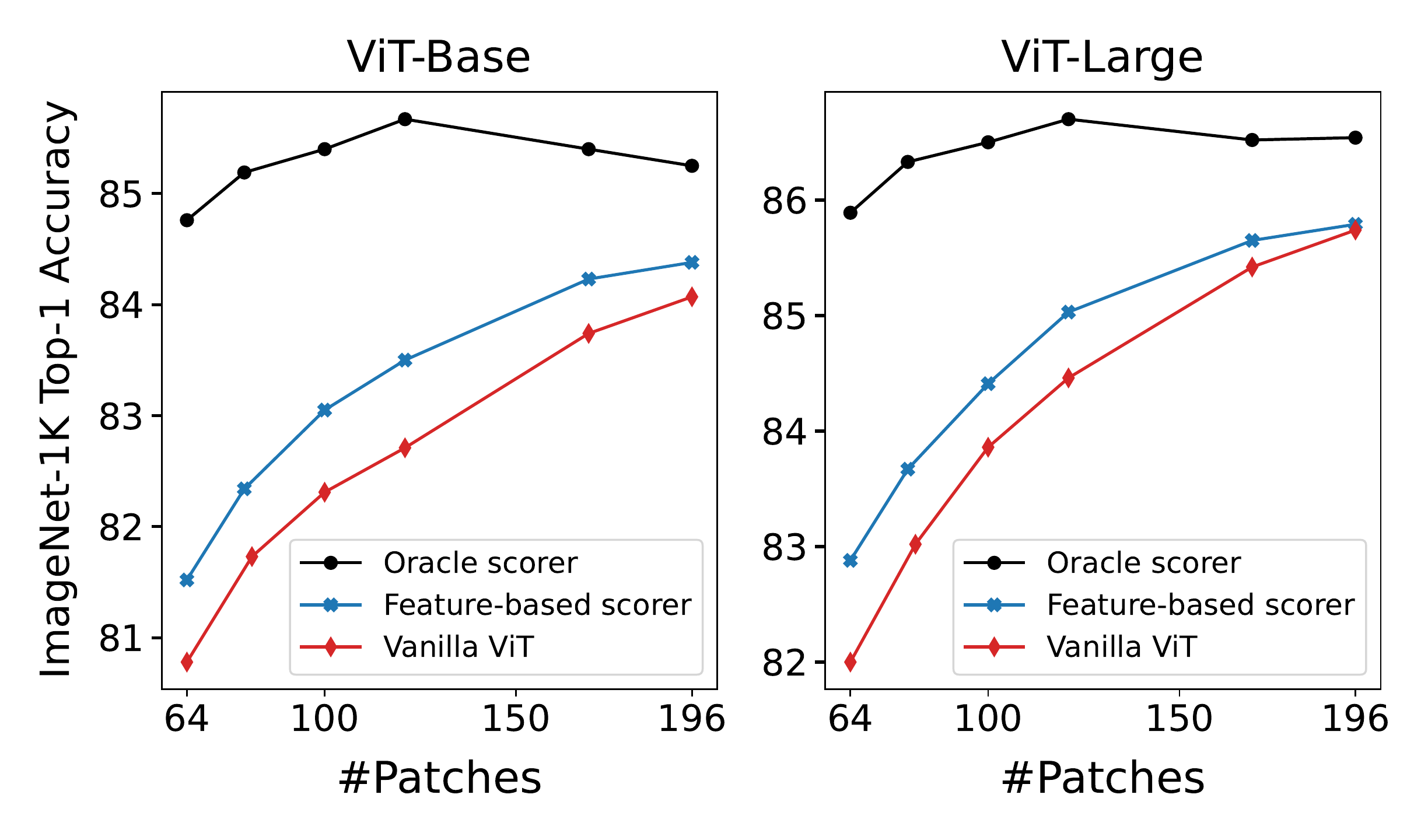}
\caption{Quadformers with a Grad-CAM oracle scorer greatly surpass vanilla ViT models, suggesting there is considerable redundancy in standard ViT tokenization.}
\label{figure:oracle}
\vspace*{10pt}
\end{figure}

\section{Analysis}

\subsection{Oracle Quadformers}
To obtain a loose upper bound on the potential performance of ViTs with mixed-resolution tokenization, we train Quadformer models with a Grad-CAM oracle saliency scorer that has access to the true image label (\S\ref{paragraph:oracle_scorer}). Our oracle models greatly surpass the performance of vanilla ViT models with the same number of patches -- in some cases by about 4 absolute percentage points (Figure \ref{figure:oracle}). Oracle Quadformers with 64 patches even beat vanilla ViTs with 196 patches despite using $\times 3$ less patches, suggesting there is considerable redundancy in standard ViT tokenization.

\subsection{Runtime breakdown}
The Transformer model and the saliency-based Quadtree tokenizer have very different runtime-to-GMACs ratios due to the different operations they use, with the tokenizer using a tiny number of GMACs compared to its runtime (Table \ref{table:cost_breakdown}). Therefore, we find that measuring actual runtime instead of settling for GMACs as the sole cost indicator is especially important when comparing our Quadformer models to vanilla ViT models.

The feature-based scorer requires 3 forward passes with a truncated ShuffleNetV2$\times0.5$, composed mainly of group convolutions, depthwise convolutions, BatchNorms, and a channel-shuffle operation that has no GMAC cost but has a non-negligible time cost.

The Quadtree algorithm itself is very fast, though it has an especially high runtime-to-GMACs ratio, as it mostly requires indexing and reshaping operations that have no GMAC cost. Even though different numbers of patches require different numbers of splits, the bulk of the Quadtree runtime is spent preparing the input to the splitting phase and processing its output, making the Quadtree cost almost constant with respect to the number of patches.

The Transformer model is composed mainly of Attention layers, fully-connected layers and LayerNorms. Its runtime and GMAC cost depend heavily on the number of patches and the size of the model.

\begin{figure}[t!]
  \centering
  \vspace*{-5pt}
  \hspace*{-0.035\linewidth}
  \includegraphics[width=1.035\linewidth]{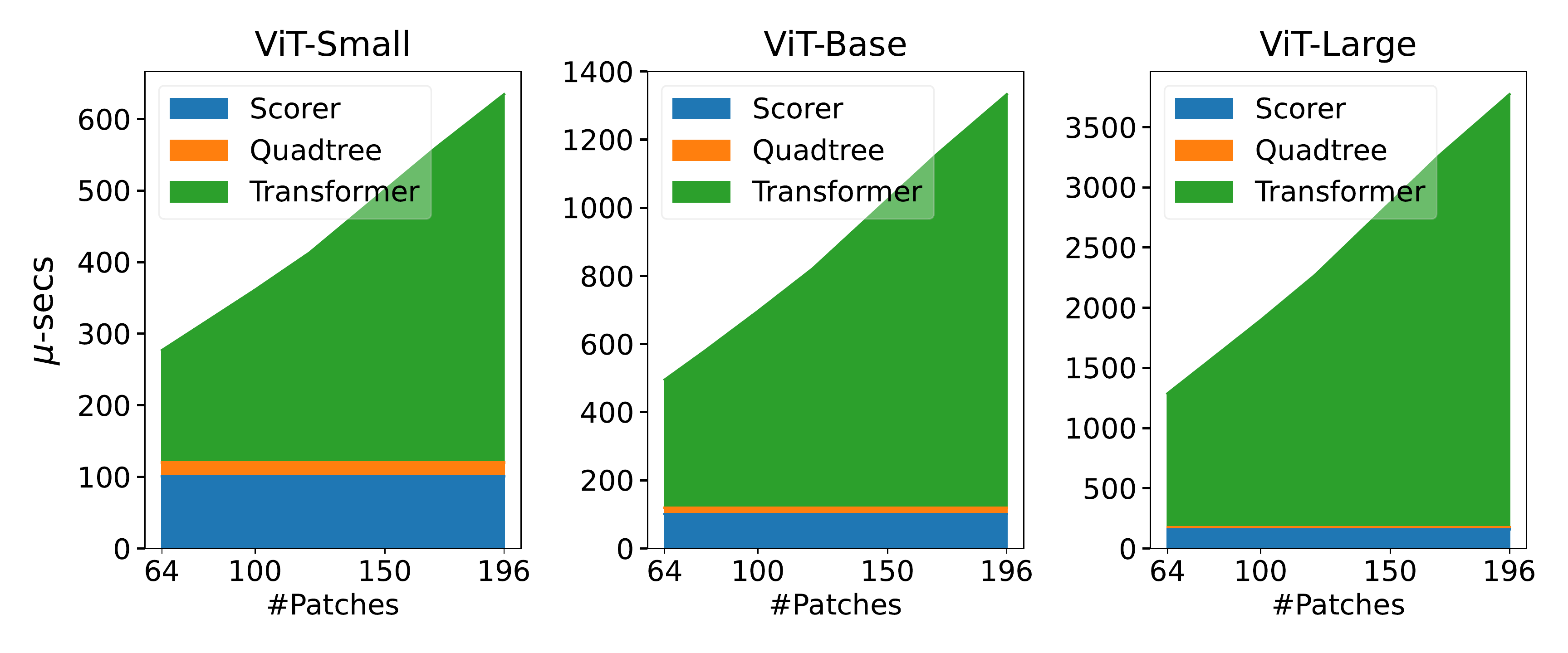}
  \caption{Quadformer forward pass runtime breakdown with a feature-based patch scorer. The fraction of time spent on tokenization changes drastically as the model and input length increase in size, from 44\% for ViT-Small with 64 patches to 4\% for ViT-Large with 196 patches.}
  \label{figure:runtime_breakdown}
\end{figure}

Notice that the fraction of time taken by patch scoring and Quadtree calculation becomes less and less significant as the model and input length increase in size (Figure \ref{figure:runtime_breakdown}), ranging from 44\% for our lightest configuration to 4\% for our heaviest.

\begin{table}[t!]
\centering
\resizebox{1.0\linewidth}{!}{
\begingroup
\renewcommand{\arraystretch}{1.2}
\begin{tabular}{ c l c c c }
Component             &                             & $\mu$-secs     & GMACs      &    $\frac{\mu-secs}{GMAC}$   \\
\midrule
Quadtree              &                             & 19             & 0.0008     &     23,750                    \\
\midrule
                      &   Pixel-Blur                & 19             & 0.0024     &     7917                     \\
Patch Scorer          &   Feature-Based $256^2$     & 157            & 0.166      &     945                      \\
                      &   Feature-Based $192^2$     & 101            & 0.094      &     1074                      \\
\midrule
                      &   ViT-Small 64 patches      & 154            & 1.44       &     107                        \\
Transformer           &   ViT-Base 121 patches     & 821            & 10.8        &     76                        \\
                      &   ViT-Large 196 patches     & 3774           & 61.8       &     61                         \\
\midrule
\end{tabular}
\endgroup
}
\caption{Forward pass cost breakdown. Measuring actual runtime is especially important for comparison between Quadformers and vanilla ViTs since different components have very different runtime-to-GMACs ratios.}
\label{table:cost_breakdown}
\end{table}

\subsection{Patch scorer quality}
The main way in which we assess the quality of different patch scorers is by measuring their effect on the downstream task, ImageNet-1K classification. Alternatively, we can measure scoring quality more directly by comparing the patch ranking induced by a Grad-CAM oracle scorer to the rankings induced by different realistic patch scorers (Figure \ref{figure:importance_map_comparison}). The oracle scorer is aware of the true image label, and uses the Grad-CAM algorithm which was built with the express purpose of saliency estimation, making it a good golden standard for patch ranking, as reflected in the high accuracy of oracle-based Quadformers (Figure \ref{figure:oracle}).

For each image in the ImageNet-1K validation dataset, we calculate rank correlation coefficients between the oracle scores and the scores computed by the feature-based and pixel-blur patch scorers. We use rank correlations since the actual score values do not affect the Quadtree algorithm, only the relative ranking (see the $\arg\max$ operation in Algorithm \ref{algorithm:quadtree}). In Table \ref{table:scorer_similarity_to_oracle} we report average rank correlation values and the fraction of images in the dataset for which the feature-based scorer was a better estimator of the oracle than the pixel-blur scorer, demonstrating the superiority of semantic representations over surface details.

\subsection{Quadtree composition}
\begin{figure}[t!]
  \centering
  \hspace*{-0.035\linewidth}
  \includegraphics[width=1.07\linewidth]{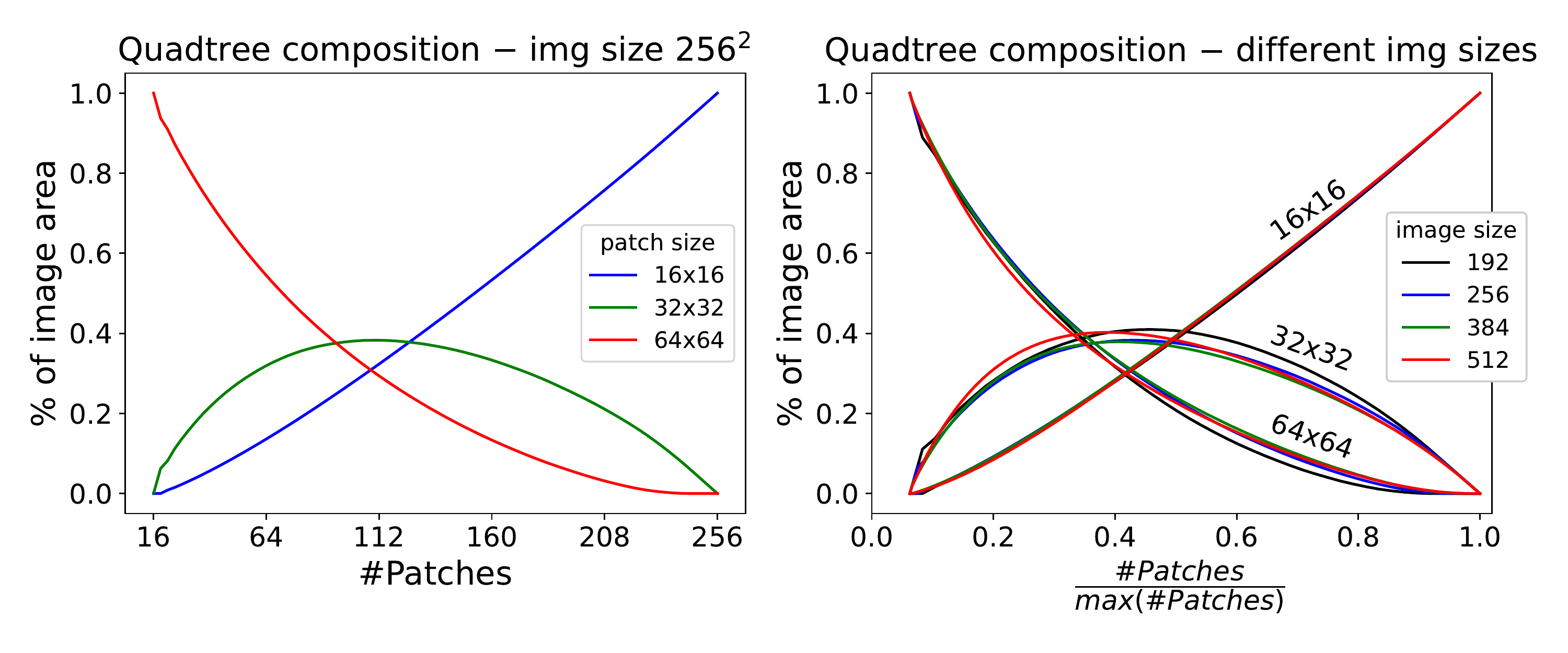}
  \caption{Quadtree composition changes with the number of splits. We measure the percentage of image area covered by each patch size to get a sense of the resolution distribution inside the image. The left plot shows the progression for our main image size. The right plot shows that this progression is almost invariant to the image size.}
  \label{figure:patch_stats}
\end{figure}

Quadtree composition changes with the number of splits. As the iterative splitting process progresses, large patches are split into medium patches, which are in turn split into small patches. While the exact frequency of different patch sizes depends on the image content, we can get a sense of the resolution distribution by constructing Quadtrees over the entire ImageNet-1K validation set and measuring the average percentage of image area covered by each patch size (Figure \ref{figure:patch_stats}). Note that the average resolution distribution depends on the ratio $\frac{\#Patches}{max(\#Patches)}$ and is almost invariant to the image size, which can help choose appropriate values for $\#Patches$ for different datasets, depending on the fraction of key information we expect the images to contain.

\begin{table}[t!]
\centering
\hspace*{-0.035\linewidth}
\resizebox{1.045\linewidth}{!}{
\begingroup
\renewcommand{\arraystretch}{1.2}
\begin{tabular}{ c c c c }
                                &  \multicolumn{2}{c}{Similarity to Oracle}    &         \\
Rank Correlation Coefficient    &  Feature-Based   & Pixel-Blur   &  \% Score$_{Feat}$ better   \\
\midrule
Kendall's $\tau$                &  0.51            & 0.31         & 81\%  \\
\midrule
Spearman                        &  0.52            & 0.26         & 81\%  \\
\midrule
\vspace*{0.001pt}
\end{tabular}
\endgroup
}
\caption{Average rank correlation between the patch rankings induced by the Grad-CAM oracle scorer and different realistic patch scorers, computed over the ImageNet-1K validation set.  ``\%~Score$_{Feat}$ better'' measures how frequently the oracle ranking is closer to the feature-based scorer than to the pixel-blur scorer.}
\label{table:scorer_similarity_to_oracle}
\end{table}

\section{Conclusion}
We have presented a novel tokenization scheme for Vision Transformers, replacing the standard uniform patch grid with a mixed-resolution sequence of tokens, where each token represents a patch of arbitrary size. We integrated the Quadtree algorithm with a novel feature-based saliency scorer to create mixed-resolution patch mosaics, making this work the first to use the Quadtree representations of RGB images as inputs for a neural network.

Through experiments in image classification, we have shown the capacity of standard Vision Transformer models to adapt to mixed-resolution tokenization via fine-tuning. Our Quadformer models achieve substantial accuracy gains compared to vanilla ViTs when controlling for the number of patches or GMACs. Although we do not use dedicated tools for accelerated inference, Quadformers also show gains when controlling for inference speed.

We believe that future work could successfully apply mixed-resolution ViTs to other computer vision tasks, especially those that involve large images with heterogeneous information densities, including tasks with dense outputs such as image generation and segmentation.

\clearpage

{\small
\bibliographystyle{ieee_fullname}
\bibliography{egbib}
}

\clearpage

\appendix

\renewcommand{\thetable}{A\arabic{table}}
\setcounter{table}{0}

\begin{center}
\LARGE
\textbf{Supplementary Material}
\end{center}

\hfill \break

\section{Full results}
We report ImageNet-1k top-1 accuracy and various cost indicators for every model configuration that appears in the figures of the main text (see Table \ref{table:results_vit_small}, Table \ref{table:results_vit_base}, Table \ref{table:results_vit_large}). Throughput is measured on a single GeForce RTX 3090 GPU in mixed precision.

\section{More implementation details}
\paragraph{Hyperparameters.} We train all of our models using the timm library~\cite{rw2019timm} with the following hyperparameters: learning rate warmup for 5 epochs, learning rate cooldown for 10 epochs, cosine learning rate scheduler~\cite{Loshchilov2016SGDRSG}, weight decay 0.025, DropPath~\cite{Huang2016DeepNW} rate 0.1, AdamW~\cite{Loshchilov2017DecoupledWD} optimizer with epsilon $1\text{e-}8$, AutoAugment~\cite{Cubuk2018AutoAugmentLA} image augmentations with configuration \verb|rand-m9-mstd0.5-inc1|, mixup~\cite{Zhang2017mixupBE} alpha 0.8, cutmix~\cite{Yun2019CutMixRS} alpha 1.0, label smoothing 0.1. Unless otherwise specified, we use base learning rate $5\text{e-}5$.

We fine-tune ViT-Small models for 130 epochs with batch size 1024, ViT-Base models for 60 epochs with batch size 400, and ViT-Large models for 20 epochs with batch size 192. For evaluation, we use exponential moving average (EMA)~\cite{Polyak1992AccelerationOS} with decay 0.99996. We use the default values in timm for all other hyperparameters.

\hfill \break

\begin{table}[h!]
\centering
\textbf{ViT-Small}
\resizebox{0.95\linewidth}{!}{
\begingroup
\renewcommand{\arraystretch}{1.1}
\begin{tabular}{ | c | c c c c | c | }
\hline
\multirow{2}{*}{Method}   &   \multirow{2}{*}{\#Patches}  &  \multirow{2}{*}{GMACs}   &  Throughput  & Runtime  & ImageNet-1k   \\
   &  &  & ims/sec   & $\mu$-secs/im   & Top-1 Acc.   \\
\hline
\multirow{5}{*}{Vanilla ViT}    & 64    & 1.44   & 6489   & 154   & 74.55   \\
 & 81    & 1.83   & 5208   & 192   & 76.36   \\
 & 100   & 2.28   & 4212   & 237   & 77.55   \\
 & 121   & 2.78   & 3460   & 289   & 78.26   \\
 & 169   & 3.94   & 2315   & 432   & 79.84   \\
 & 196   & 4.62   & 1975   & 506   & 80.28   \\
\hline
\multirow{5}{*}{\shortstack[c]{Quadformer \\ \small{Feature-based scorer}}}   & 64    & 1.54   & 3611   & 277   & 76.53   \\
 & 79    & 1.88   & 3204   & 312   & 77.53   \\
 & 100   & 2.37   & 2766   & 362   & 78.64   \\
 & 121   & 2.87   & 2419   & 413   & 79.35   \\
 & 169   & 4.04   & 1792   & 558   & 80.43   \\
 & 196   & 4.71   & 1576   & 635   & 80.84   \\
\hline
\multirow{5}{*}{\shortstack[c]{Quadformer \\ \small{Pixel-blur scorer}}}    & 64    & 1.45   & 5150   & 194   & 74.97   \\
 & 79    & 1.79   & 4362   & 229   & 76.27   \\
 & 100   & 2.28   & 3590   & 279   & 77.47   \\
 & 121   & 2.78   & 3022   & 331   & 78.58   \\
 & 169   & 3.95   & 2104   & 475   & 80.01   \\
 & 196   & 4.62   & 1813   & 552   & 80.4   \\
\hline
\end{tabular}
\endgroup
}
\vspace*{0.2cm}
\caption{Full results - ViT Small.}
\label{table:results_vit_small}
\end{table}

\hfill \break

\hfill \break

\hfill \break

\hfill \break

\begin{table}[h!]
\centering
\textbf{ViT-Base}
\resizebox{0.95\linewidth}{!}{
\begingroup
\renewcommand{\arraystretch}{1.1}
\begin{tabular}{ | c | c c c c | c | }
\hline
\multirow{2}{*}{Method}   &   \multirow{2}{*}{\#Patches}  &  \multirow{2}{*}{GMACs}   &  Throughput  & Runtime  & ImageNet-1k   \\
   &  &  & ims/sec   & $\mu$-secs/im   & Top-1 Acc.   \\
\hline
\multirow{5}{*}{Vanilla ViT}   & 64  &  5.6 & 2676   & 374   & 80.78   \\
   & 81  &  7.2  & 2155   & 464    & 81.73   \\
   & 100 &  8.8  & 1739   & 575    & 82.31   \\
   & 121 &  10.7  & 1429   & 700    & 82.71   \\
   & 169 &  15.1  & 966    & 1035   & 83.74   \\
   & 196 &  17.6  & 823    & 1215   & 84.07   \\
\hline
\multirow{5}{*}{\shortstack[c]{Quadformer \\ \small{Feature-based scorer}}}   &   64    & 5.7  &   2019   &   495   &   81.52   \\
   &   79   & 7.1  &   1732   &   577   &   82.34   \\
   &   100  & 8.9  &   1435   &   697   &   83.05   \\
   &   121  & 10.8 &   1218   &   821   &   83.50   \\
   &   169  & 15.2 &   864    &   1157  &   84.23   \\
   &   196  & 17.7 &   750    &   1333  &   84.38   \\
\hline
\multirow{5}{*}{\shortstack[c]{Quadformer \\ \small{Pixel-blur scorer}}}   & 64 &  5.7  & 2424   & 413   & 80.78   \\
   & 79  & 7.0  & 2021   & 495   & 81.68  \\
   & 100 & 8.8  & 1630   & 613   & 82.57  \\
   & 121 & 10.7 & 1354   & 739   & 83.06   \\
   & 169 & 15.1 & 931    & 1074  & 83.87   \\
   & 196 & 17.6 & 800    & 1250  & 84.23   \\
\hline
\multirow{5}{*}{\shortstack[c]{Quadformer \\ \small{Oracle scorer}}}   & 64   &  ---   &  ---   & ---   & 84.76   \\
   & 79    &  ---   &  ---   & ---   & 85.19   \\
   & 100   &  ---   &  ---   & ---   & 85.40   \\
   & 121   &  ---   &  ---   & ---   & 85.67   \\
   & 169   &  ---   &  ---   & ---   & 85.40   \\
   & 196   &  ---   &  ---   & ---   & 85.25   \\
\hline
\end{tabular}
\endgroup
}
\vspace*{0.2cm}
\caption{Full results - ViT Base.}
\label{table:results_vit_base}
\end{table}

\begin{table}[h!]
\centering
\textbf{ViT-Large}
\resizebox{0.95\linewidth}{!}{
\begingroup
\renewcommand{\arraystretch}{1.1}
\begin{tabular}{ | c | c c c c | c | }
\hline
\multirow{2}{*}{Method}   &   \multirow{2}{*}{\#Patches}  &  \multirow{2}{*}{GMACs}   &  Throughput  & Runtime  & ImageNet-1k   \\
   &  &  & ims/sec   & $\mu$-secs/im   & Top-1 Acc.   \\
\hline
\multirow{5}{*}{Vanilla ViT}   & 64 &  19.9   & 900   & 1111   & 82.00   \\
   & 81  & 25.2   & 720   & 1389   & 83.02   \\
   & 100 & 31.1   & 580   & 1724   & 83.86   \\
   & 121 & 37.7   & 478   & 2092   & 84.46   \\
   & 169 & 53.0   & 323   & 3096   & 85.42   \\
   & 196 & 61.7   & 277   & 3610   & 85.74   \\
\hline
\multirow{5}{*}{\shortstack[c]{Quadformer \\ \small{Feature-based scorer}}}   & 64   & 20.1   & 777   & 1287   & 82.88   \\
   & 79  &  24.7  & 649   & 1541   & 83.67   \\
   & 100 &  31.3  & 527   & 1898   & 84.41   \\
   & 121 &  37.9  & 440   & 2273   & 85.03   \\
   & 169 &  53.1  & 306   & 3268   & 85.65   \\
   & 196 &  61.8  & 265   & 3774   & 85.79   \\
\hline
\multirow{5}{*}{\shortstack[c]{Quadformer \\ \small{Pixel-blur scorer}}}  & 64  & 19.9   & 869   & 1151  & 81.66  \\
 & 79  & 24.6   & 712   & 1404  & 82.69  \\
 & 100 & 31.1   & 568   & 1761  & 83.61  \\
 & 121 & 37.7   & 470   & 2128  & 84.3   \\
 & 169 & 53.0   & 320   & 3125  & 85.22  \\
 & 196 & 61.7   & 275   & 3636  & 85.56  \\
\hline
\multirow{5}{*}{\shortstack[c]{Quadformer \\ \small{Oracle scorer}}}   & 64   &  ---   &  ---   & ---   & 85.89   \\
   & 79    &  ---   &  ---   & ---   & 86.33   \\
   & 100   &  ---   &  ---   & ---   & 86.5   \\
   & 121   &  ---   &  ---   & ---   & 86.7   \\
   & 169   &  ---   &  ---   & ---   & 86.52   \\
   & 196   &  ---   &  ---   & ---   & 86.54   \\
\hline
\end{tabular}
\endgroup
}
\vspace*{0.2cm}
\caption{Full results - ViT-Large.}
\label{table:results_vit_large}
\end{table}

\end{document}